\documentclass[runningheads]{llncs}
\usepackage[dvipdfmx]{graphicx}
\usepackage{amsmath,amssymb} 
\usepackage{color}
\usepackage{upgreek}
\usepackage{mathtools}
\usepackage{booktabs, siunitx} 
\usepackage[width=122mm,left=12mm,paperwidth=146mm,height=193mm,top=12mm,paperheight=217mm]{geometry}
\usepackage{comment}

\def\etal{\emph{et al. }}
\newcommand{\Tref}[1]{Table~\ref{#1}}
\newcommand{\Eref}[1]{Eq.~\ref{#1}}
\newcommand{\Fref}[1]{Fig.~\ref{#1}}
\newcommand{\Sref}[1]{Sec.~\ref{#1}}
\newcommand{\bvec}[1]{\mathbf{#1}}
\newcommand{\simthree}{\mathrm{Sim}(3)}
\newcommand{\sethree}{\mathrm{SE}(3)}
\newcommand{\sothree}{\mathrm{SO}(3)}

\begin{document}

\pagestyle{headings}
\mainmatter

\title{Scale Drift Correction of Camera Geo-Localization using Geo-Tagged Images} 

\titlerunning{Scale Drift Correction of Camera Geo-Localization using Geo-Tagged Images}

\authorrunning{K. Iwami and S. Ikehata and K. Aizawa}

\author{Kazuya Iwami\inst{1}  and Satoshi Ikehata\inst{2}  and Kiyoharu Aizawa\inst{1} }
\institute{The University of Tokyo, Japan\\
\email{\{iwami,aizawa\}@hal.t.u-tokyo.ac.jp}\\
\and
National Institute of Informatics, Japan\\
\email{sikehata@nii.ac.jp}}

\maketitle

\begin{abstract}
Camera geo-localization from a monocular video is a fundamental task for video analysis and autonomous navigation. Although 3D reconstruction is a key technique to obtain camera poses, monocular 3D reconstruction in a large environment tends to result in the accumulation of errors in rotation, translation, and especially in scale: a problem known as scale drift. To overcome these errors, we propose a novel framework that integrates incremental structure from motion (SfM) and a scale drift correction method utilizing geo-tagged images, such as those provided by Google Street View. Our correction method begins by obtaining sparse 6-DoF correspondences between the reconstructed 3D map coordinate system and the world coordinate system, by using geo-tagged images. Then, it corrects scale drift by applying pose graph optimization over $\simthree$ constraints and bundle adjustment. Experimental evaluations on large-scale datasets show that the proposed framework not only sufficiently corrects scale drift, but also achieves accurate geo-localization in a kilometer-scale environment.

\end{abstract}

\section{Introduction}

Camera geo-localization from a monocular video in a kilometer-scale environment is a essential technology for AR, video analysis, and autonomous navigation. To achieve accurate geo-localization, 3D reconstruction from a video is a key technique. Incremental structure from motion (SfM) and visual simultaneous localization and mapping (visual SLAM) achieve large-scale 3D reconstructions by simultaneously localizing camera poses with six degrees-of-freedom (6-DoF) and reconstructing a 3D environment map~\cite{engel2014lsd,mur2015orb}.

Unlike for a stereo camera, an absolute scale of the real world cannot be derived using a single observation from a monocular camera. Although it is possible to estimate an environment's relative scale from a series of monocular observations, errors in the relative scale estimation accumulate over time, and this is referred to as scale drift~\cite{clemente2007mapping,strasdat2010scale}. 

For an accurate geo-localization not affected by scale drift ,prior information in a geographic information system (GIS)  has been utilized in previous studies. For example, point clouds, 3D models, building footprints, and road maps have been proven to be efficient for correcting reconstructed 3D maps~\cite{middelberg2014scalable,caselitz2016monocular,tamaazousti2011nonlinear,untzelmann2013scalable,brubaker2016map}. However, these priors are only available in limited situations, e.g., in an area that is observed in advance, or in an environment consisting of simply-shaped buildings. Therefore, there is a good chance that other GIS information can help to extend the area in which a 3D map can be corrected.

Hence, in this paper, motivated by the recent availability of massive public repositories of geo-tagged images taken all over the world, we propose a novel framework for correcting the scale drift of monocular 3D reconstruction by utilizing geo-tagged images, such as those in Google Street View~\cite{street_view}, and achieve accurate camera geo-localization.
Owing to the high coverage of Google Street View, our proposal is more scalable than those in previous studies.

\begin{figure}[t]
	\centering
		\includegraphics[width=0.95\linewidth]{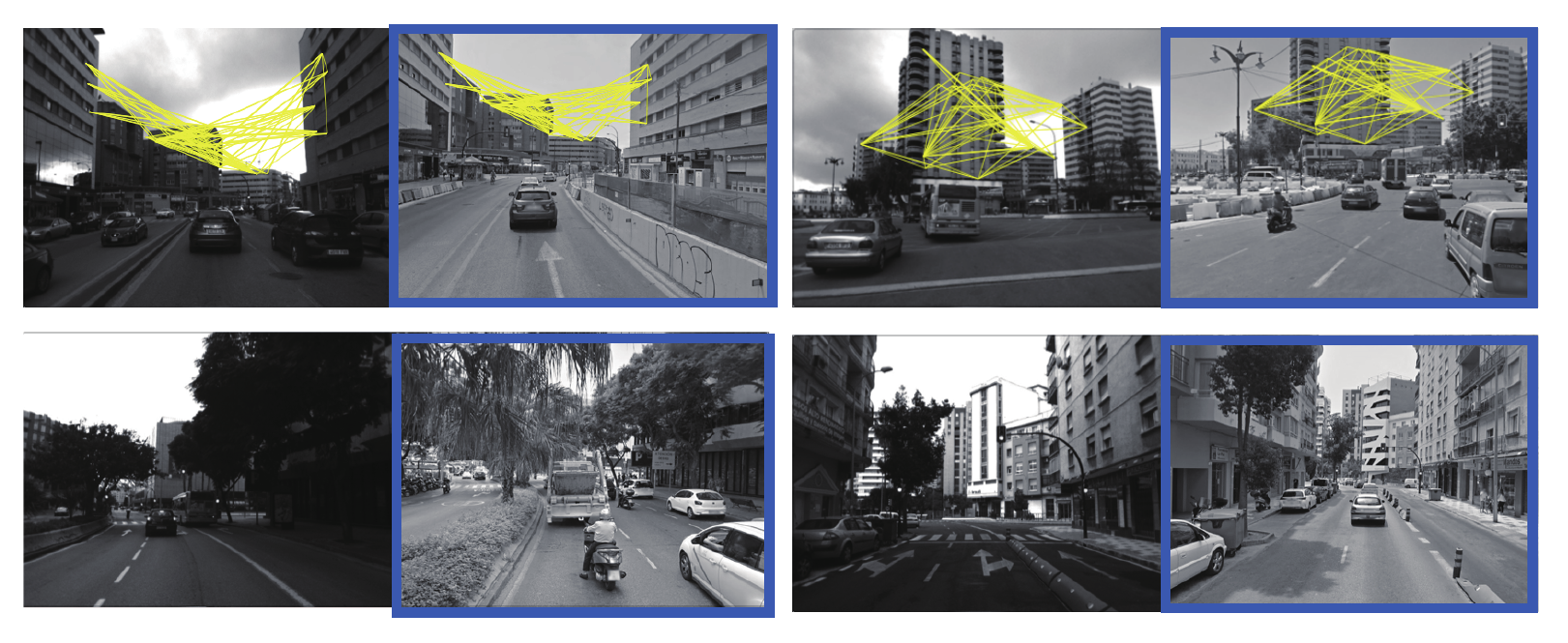}
		\caption{Examples of keypoint matches between keyframes (without blue squares) and geo-tagged images of Google Street View (with blue squares) after kVLD validation. Top: pairs of images where valid matches are found. Yellow lines denote kVLD graph structures, which are composed of inliers. Bottom: rejected pairs of images where a sufficient number of matches is not found because of differences in illumination, viewpoint, and environment, despite being taken in almost the same location.}
		\label{fig:correspondence}
\end{figure}

The proposed framework integrates incremental SfM and a scale drift correction method utilizing geo-tagged images. Our correction method begins by computing 6-DoF correspondences between the reconstructed 3D map coordinate system and the world coordinate system, by using geo-tagged images. Owing to significant differences in illumination, viewpoint, and the environment resulting from differences in time, it tends to be difficult to acquire correspondences between video frames and geo-tagged images (\Fref{fig:correspondence}). Therefore, a new correction method that can deal with the large scale drift of a 3D map using a limited number of correspondences is required. Bundle adjustment with constraints of global position information, which represents one of the most important correction methods, cannot be applied directly. This is because bundle adjustment tends to get stuck in a local minimum when starting from a 3D map including large errors~\cite{strasdat2010scale}. Hence, the proposed correction method consists of two coarse-to-fine steps: pose graph optimization over Sim(3) constraints, and bundle adjustment. In these steps, our key idea is to extend the pose graph optimization method proposed for the loop closure technique of monocular SLAM~\cite{strasdat2010scale}, such that it incorporates the correspondences between the 3D map coordinate system and the world coordinate system. This step corrects the large errors, and enables bundle adjustment to obtain precise results. After implementing this framework, we conducted experiments to evaluate the proposal.

The contributions of this work are as follows. First, we propose a novel framework for camera geo-localization that can correct scale drift by utilizing geo-tagged images. Second, we extend the pose graph optimization approach to dealing with scale drift using a limited number of correspondences to geo-tags. Finally, we validate the effectiveness of the proposal through experimental evaluations on kilometer-scale datasets.

\section{Related Work}
\subsection{Monocular 3D Reconstruction\label{sec:related_work_correction}}
Incremental SfM and visual SLAM are important approaches to reconstructing 3D maps from monocular videos. Klein~\etal proposed PTAM for small AR workspaces~\cite{klein2007parallel}. Mur-Artal~\etal developed ORB-SLAM, which can reconstruct large-scale outdoor environments~\cite{mur2015orb}. For accurate 3D reconstruction, the loop closure technique has commonly been employed in recent SLAM approaches~\cite{strasdat2010scale,mur2015orb}. Loop closure deals with errors that accumulate between two camera poses that occur at the same location, i.e., when the camera trajectory forms a loop. Lu and Milios~\cite{lu1997globally} formulated this technique as a pose graph optimization problem, and Strasdat~\etal~\cite{strasdat2010scale} extended pose graph optimization to deal with scale drift for monocular visual SLAM. It is certain that loop closure can significantly improve 3D maps, but this is only effective if a loop exists in the video.

\subsection{Geo-registration of Reconstructions}
Correcting reconstructed 3D maps by using geo-referenced information has been regarded as a geo-registration problem. Kaminsky~\etal proposed a method that aligns 3D reconstructions to 2D aerial images~\cite{kaminsky2009alignment}. Wendel~\etal used an overhead digital surface model (DSM) for the geo-registration of 3D maps~\cite{wendel2011automatic}. Similar to our work, Wang~\etal used Google Street View geo-tagged images and a Google Earth 3D model for the geo-registration of reconstructed 3D maps~\cite{wang2013accurate}. However, because all these methods focus on estimating a best-fitting similarity transformation to geo-referenced information, they only correct the global scale in terms of 3D map correction.

Methods for geo-registration using non-linear transformations have also been proposed. 
To integrate GPS information, Lhuillier~\etal proposed incremental SfM using bundle adjustment with constraints from GPS~\cite{lhuillier2012incremental}, and Rehder~\etal formulated a global pose estimation problem using stereo visual odometry, inertial measurements, and infrequent GPS information as a 6-DoF pose graph optimization problem~\cite{rehder2012global}. In terms of correcting camera poses using sparse global information, Rehder's method is similar to our pose graph optimization approach. However, our 7-DoF pose graph optimization differs in focusing on scale drift resulting from monocular 3D reconstruction, and in utilizing geo-tagged images. In addition to GPS information, various kinds of reference data have been used for the non-linear geo-registration or geo-localization of a video, such as point clouds~\cite{middelberg2014scalable,caselitz2016monocular}, 3D models~\cite{tamaazousti2011nonlinear}, building footprints~\cite{untzelmann2013scalable}, and road maps~\cite{brubaker2016map}. In this paper, we address a method that introduces geo-tagged images to the non-linear geo-registration of 3D maps.

\section{Proposed Method}
\begin{figure}[tb]
	\centering
		\includegraphics[width=0.6\linewidth]{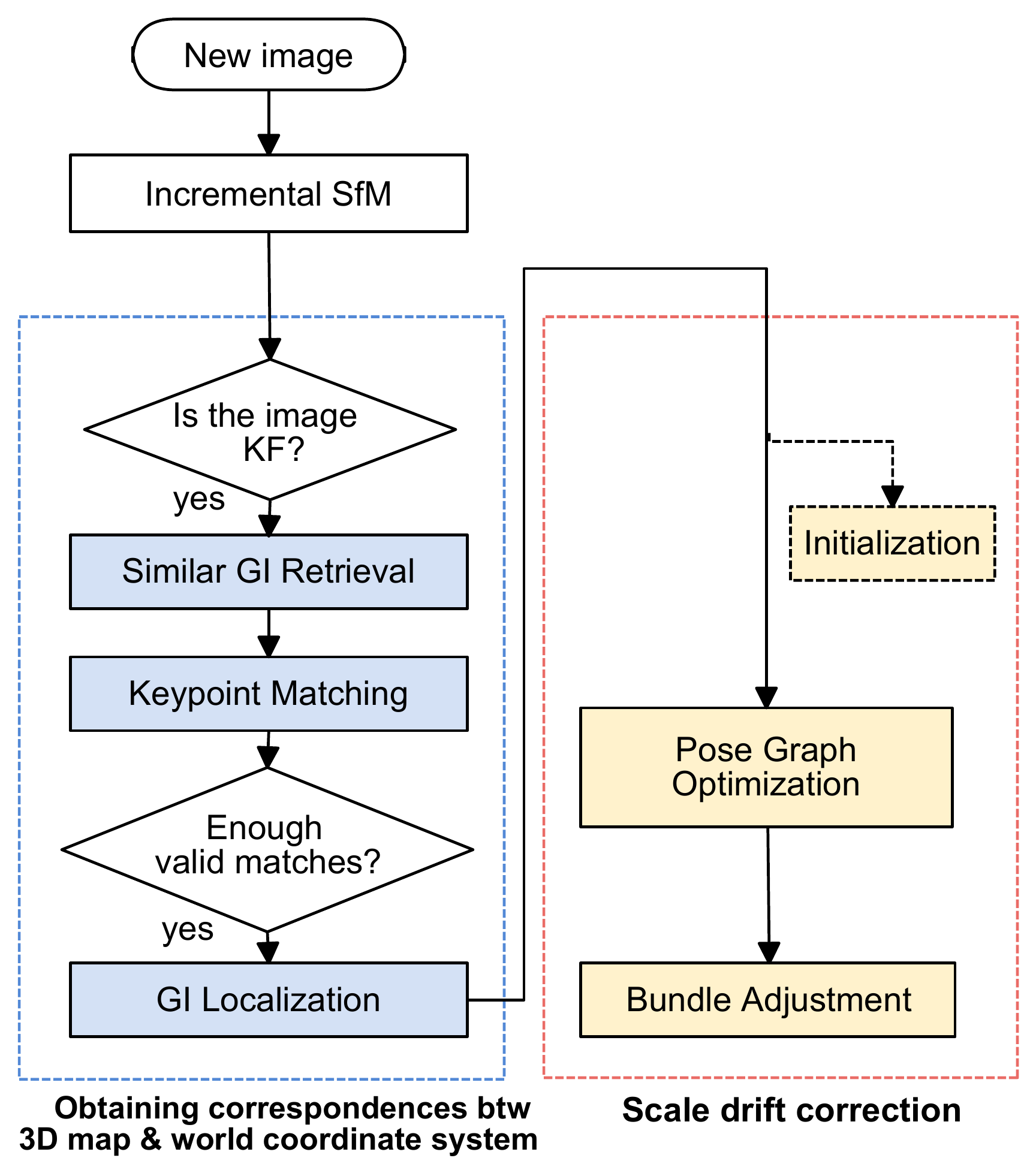}
		\caption{A flowchart of our proposal. KF and GI denote a keyframe and geo-tagged image, respectively. Initialization is performed only once in a whole reconstruction.}
		\label{fig:flowchart}
\end{figure}

\Fref{fig:flowchart} provides a flowchart of the proposed framework, which is roughly divided into three parts. The first part is incremental SfM, and is described in \Sref{sec:incremental}. The second part computes 6-DoF correspondences between the 3D map coordinate system and the world coordinate system (as defined below), by making use of geo-tagged images (\Sref{sec:get_correspondence}). The third part then uses the correspondences to correct the scale drift of the 3D map, by applying pose graph optimization over Sim(3) constraints (\Sref{sec:PGO}) and bundle adjustment (\Sref{sec:BA}) incrementally. The initialization of the scale drift correction method is described in \Sref{sec:Init}. 

\subsection{World Coordinate System}
In this paper, the world coordinates are represented by 3D coordinates $(x, y, z)$, where the $xz$-plane corresponds to the Universal Transverse Mercator (UTM) coordinate system, which is an orthogonal coordinate system using meters, and $y$ corresponds to the height from the ground in meters. The UTM coordinates can be converted into latitude and longitude if necessary.  

\subsection{Incremental SfM \label{sec:incremental}}
As large-scale incremental SfM, we use ORB-SLAM~\cite{mur2015orb} (with no real-time constraints). This is one of the best-performing monocular SLAM systems. Frames that are important for 3D reconstruction are selected as keyframes by ORB-SLAM. Every time a new keyframe is selected, our correction method is performed, and the 3D map reconstructed up to that point is corrected. In the 3D reconstruction, we identify 3D map points and their corresponding 2D keypoints in the keyframes (collectively denoted by $C_{\scalebox{0.7}{map-kf}}$).

Our proposed framework does not depend on a certain 3D reconstruction method, and can be applied to the other monocular 3D reconstruction methods, such as incremental SfM and feature-based visual SLAM.

\subsection{Obtaining Correspondences between 3D Map and World Coordinates \label{sec:get_correspondence}}
Here, we describe the second part of the proposed method, which uses geo-tagged images to compute a 6-DoF correspondence, $C_{\scalebox{0.7}{map-world}}$, between the 3D map and world coordinate system.
For this purpose, we modify Agarwal's method~\cite{agarwal2015metric} to integrate it into ORB-SLAM. This part consists of the following four steps: geo-tagged image collection, similar geo-tagged image retrieval, keypoint matching, and geo-tagged image localization.
\\
\\
\textbf{Geo-tagged Image Collection.} 
Google Street View~\cite{street_view} is a browsable street-level GIS, which is one of the largest repositories of global geo-tagged images (i.e., images and their associated geo-tags). All images are high-resolution RGB panorama images, containing highly accurate world positions~\cite{klingner2013street}. We make use of this data by converting each panorama image into eight rectilinear images with the same field-of-view as our input video, with eight horizontal directions. 
Note that because each geo-tag has a position and rotation in the world coordinates, we can obtain the 6-DoF correspondences between the 3D map coordinate system and world coordinate system if geo-tagged images are localized in the 3D map coordinate system.
\\
\\
\textbf{Similar Geo-tagged Image Retrieval.} 
When a new keyframe is selected, we retrieve the top-$k$ similar geo-tagged images. The retrieval system employs a bag-of-words approach based on SIFT descriptors \cite{agarwal2015metric}.
\\
\\
\textbf{Keypoint Matching.} 
Given the pairs of keyframes and retrieved geo-tagged images, we detect ORB keypoints~\cite{rublee2011orb} from the pairs and perform keypoint matching. Because the matching between video frames and Google Street View images tends to include many outliers~\cite{majdik2013mav}, we use a virtual line descriptor (kVLD)~\cite{liu2012virtual}, which can reject outliers by using a graph matching method even when inlier rate is around 10\
\\
\\
\textbf{Geo-tagged Image Localization.} 
To compute $C_{\scalebox{0.7}{map-world}}$, we first compute 3D-to-2D correspondences $C_{\scalebox{0.7}{map-geo}}$ between 3D map points and their corresponding 2D keypoints in geo-tagged images. In particular, we obtain $C_{\scalebox{0.7}{map-geo}}$ by combining the 2D keypoint matches (computed in the previous step) with the correspondences  $C_{\scalebox{0.7}{map-kf}}$ between 3D map points and their corresponding 2D keypoints in keyframes (computed in 3D reconstruction). 
Then, we obtain the 6-DoF camera poses of geo-tagged images in the 3D map coordinate system by minimizing the re-projection errors of $C_{\scalebox{0.7}{map-geo}}$, using the LM algorithm.
Finally, we obtain $C_{\scalebox{0.7}{map-world}}$ by combining the camera poses of geo-tagged images and 6-DoF camera poses of the associated geo-tags.

\begin{figure}[t]
	\centering
		\includegraphics[width=\linewidth]{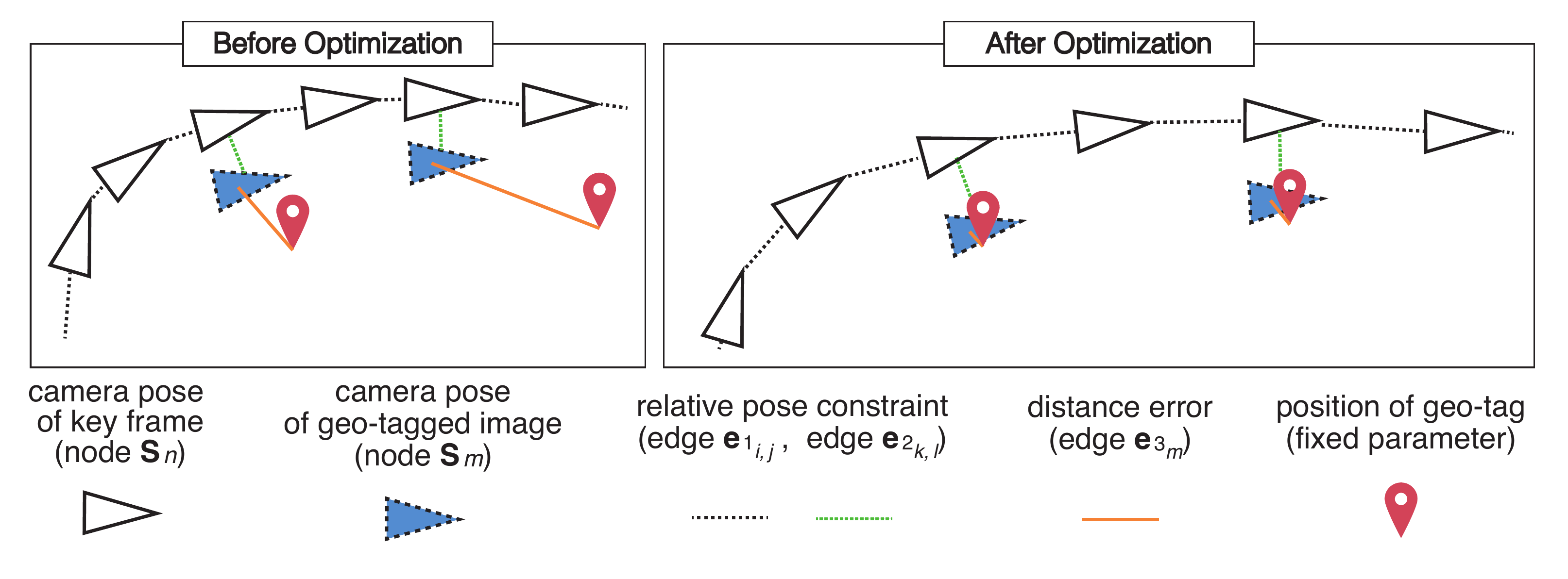}
		\caption{An example of the proposed pose graph optimization. This optimization maintains overall relative poses, except for gradual scale changes, and keeps camera poses of geo-tagged images close to the positions of the corresponding geo-tags.}
		\label{fig:pgo}
\end{figure}

\subsection{Initialization (INIT) \label{sec:Init}}
As the initialization, two kinds of linear transformations are performed on the 3D map, because the positions and scales of the 3D map coordinates and world coordinates are significantly different. Initialization is applied once, when the $i$-th geo-tagged image is localized. We set $i = 4$.

Given the first to $i$-th $C_{\scalebox{0.7}{map-world}}$, the first transformation assumes that all camera poses are approximately located in one plane, and rotates the 3D map to align that plane to the world $xz$-plane. The best-fitting plane can be estimated by a principal component analysis. 

Next, we estimate the best-fitting transformation matrix given by \Eref{eq:similarity_matrix}, which transforms a point in the 3D map coordinate system $\bvec{p}_{\scalebox{0.6}{SLAM},k}$ to be closer to a corresponding point in the world coordinate system $\bvec{p}_{world,k}$ ($\bvec{p}_{\scalebox{0.6}{SLAM},k}$ and $\bvec{p}_{world,k}$ are denoted using a homogeneous representation): 
\begin{equation}\label{eq:similarity_matrix}
\bvec{A}    =
    \begin{bmatrix}
    	s * \cos(\theta) & 0 & -s * \sin(\theta) & a\\
        0 & s & 0 & 1 \\
		s * \sin(\theta) & 0 & s * \cos(\theta) & b\\
        0 & 0 & 0 & 1
	\end{bmatrix} 
\end{equation}

Using the first to $i$-th $C_{\scalebox{0.7}{map-world}}$, we estimate the four matrix parameters $[a, b, s, \theta]$ by minimizing the following cost using RANSAC~\cite{fischler1981random} and the Levenberg-Marquart (LM) algorithm: 
\begin{equation}
E    =
    \sum_{k \in 1,2...i} \| \bvec{p}_{world,k} -  \bvec{A} \bvec{p}_{\scalebox{0.6}{SLAM},k} \|^{2}
\end{equation}
The camera poses of the geo-tagged images in $C_{\scalebox{0.7}{map-world}}$, keyframes, and 3D map point can then be transformed using the resulting matrix. 

\subsection{Pose Graph Optimization over Sim(3) Constraints (PGO)\label{sec:PGO}}
We correct the 3D map focusing on scale drift by using the newest three of $C_{\scalebox{0.7}{map-world}}$. This correction is performed every time a new $C_{\scalebox{0.7}{map-world}}$ is found after initialization. Then, we propose a graph-based non-linear optimization method (pose graph optimization) on Lie manifolds, which simultaneously corrects the scale drift and aligns the 3D map with the world coordinates.  
\\
\\
\textbf{Notation.} A 3D rigid body transformation $\bvec{G} \in \sethree$ and a 3D similarity transformation $\bvec{S} \in \simthree$ are defined by \Eref{eq:se3_sim3}, where $\bvec{R} \in \sothree$, $\bvec{t} \in \mathbb{R}^3$, and $s \in \mathbb{R}^{+}$. Here, $\sothree$, $\sethree$, and $\simthree$ are Lie groups, and $\mathfrak{so}(3)$, $\mathfrak{se}(3)$, and $\mathfrak{sim}(3)$ are their corresponding Lie algebras. A Lie group can be transformed into a Lie algebra using its exponential map, and the inverse transformation is defined by the inverse logarithm map. Each Lie algebra is represented by a vector of its coefficients. 
For example, $\mathfrak{sim}(3)$ is represented as the seven-vector $\boldsymbol{\xi} = (\omega_{1},\omega_{2},\omega_{3}, \sigma, \nu_{1}, \nu_{2}, \nu_{3})^{\mathrm{T}} = (\boldsymbol{\omega}, \sigma, \boldsymbol{\nu})^{\mathrm{T}}$, and the exponential map $\exp_{\simthree}$ and logarithm map $\log_{\simthree}$ are defined as in \Eref{eq:exp_sim} and \Eref{eq:log_sim}, respectively, where $\bvec{W}$ is a term similar to Rodriguez's formula. Further details of $\simthree$ are given in \cite{strasdat2010scale}. 
\begin{align}\label{eq:se3_sim3}
\bvec{G}
= 
\begin{bmatrix}
	\bvec{R} & \bvec{t}\\
       \bvec{0} & 1
\end{bmatrix}
&\qquad
\bvec{S}
= 
\begin{bmatrix}
	s \bvec{R} & \bvec{t}\\
       \bvec{0} & 1
\end{bmatrix}
\end{align}
\begin{equation}\label{eq:exp_sim}
\begin{split}
\exp_{\simthree}(\boldsymbol{\xi})
&=
	\begin{bmatrix}
		e^{\sigma} \exp_{\sothree}(\boldsymbol{\omega}) & \bvec{W} \boldsymbol{\nu} \\
        \bvec{0} & 1
	\end{bmatrix} = \bvec{S}
\end{split}
\end{equation}
\begin{equation}\label{eq:log_sim}
\log_{\simthree}(\bvec{S}) = {\exp_{\simthree}}^{-1}(\bvec{S}) = \boldsymbol{\xi}
\end{equation}
\textbf{Proposed pose graph optimization.}
In a general pose graph optimization approach~\cite{lu1997globally,rehder2012global}, camera poses and relative transformations between two camera poses are represented as elements of $\sethree$. However, in our approach, 6-DoF camera poses and relative transformations are converted into 7-DoF camera poses, represented by elements of $\simthree$. This is achieved by leaving the rotation $R$ and translation $\bvec{t}$ of a camera pose unchanged, and setting the scale $s$ to $1$. The idea that camera poses and relative pose constraints can be handled in $\simthree$ was proposed by Strasdat~\etal~\cite{strasdat2010scale}, for dealing with the scale drift problem in monocular SLAM. In this paper, we introduce 7-DoF pose graph optimization, which has previously only been used in the context of loop closure, to correct 3D reconstruction by utilizing sparse correspondences between two coordinate systems. Our pose graph contains two kinds of nodes and three kinds of edges, as follows (see \Fref{fig:pgo}):
\begin{itemize}
	\setlength{\itemsep}{2pt}      
	\setlength{\parskip}{6pt}      
	\setlength{\itemindent}{0pt}   
    \item Node $\bvec{S}_{n} \in \simthree$, where $n \in C_{1} $: the camera pose of the $n^{th}$ keyframe.
	\item Node $\bvec{S}_{m} \in \simthree$, where $m \in C_{2}$: the camera pose of the $m^{th}$ geo-tagged image.
    \item Edge $\bvec{e}_{1_{i, j}}$, where $(i, j) \in C_{3}$: the relative pose constraint between the $i^{th}$ and $j^{th}$ keyframes. (\Eref{eq:edge1})
    \item Edge $\textbf{e}_{2_{k,l}}$, where $(k, l) \in C_{4}$: the relative pose constraint between the $k^{th}$ keyframe and the $l^{th}$ geo-tagged image. (\Eref{eq:edge2})
	\item Edge $\textbf{e}_{3_{m}}$, where $m \in C_{2}$: the distance error between the position of the $m^{th}$ geo-tagged image and the world position $\bvec{y}_{m}$ of the corresponding geo-tag. (\Eref{eq:edge3})
\end{itemize}
\begin{gather}
\label{eq:edge1}
\textbf{e}_{1_{i,j}} = \log_{\simthree}(\Delta \bvec{S}_{i,j} \cdot \bvec
{S}_{i} \cdot \bvec{S}_{j}^{-1}) \in \mathbb{R}^7 \\
\label{eq:edge2}
\textbf{e}_{2_{k,l}} =  \log_{\simthree}(\Delta \bvec{S}_{k,l} \cdot \bvec{S}_{k} \cdot \bvec{S}_{l}^{-1}) \in \mathbb{R}^7\\
\label{eq:edge3}
\textbf{e}_{3_{m}}  = \mathrm{trans}(\bvec{S}_{m}) - \bvec{y}_{m} \in \mathbb{R}^3 
\end{gather}
where $\mathrm{trans}(\bvec{S}) \equiv (\bvec{S}_{1,4}, \bvec{S}_{2,4}, \bvec{S}_{3,4})^{\mathrm{T}}$. Here, $N$ is the total number of keyframes, and $M$ is the total number of geo-tagged images that have correspondences to keyframes. The set $C_{1}$ contains all the keyframes positioned between the two that have the newest and the third newest $C_{\scalebox{0.7}{map-world}}$.  The set $C_{2}$ contains the newest three of $C_{\scalebox{0.7}{map-world}}$. The set $C_{3}$ contains the pairs of keyframes that observe the same 3D map point in 3D reconstruction, and $C_{4}$ contains pairs of keyframes and their corresponding geo-tagged images. Finally, $\Delta \bvec{S}_{i,j}$ is the converted $\simthree$ relative transformation between $\bvec{S}_{i}$ and $\bvec{S}_{j}$, which is calculated before the optimization and remains fixed during the optimization. 

Note that we newly introduced the nodes $\bvec{S}_{m}$, edges $\textbf{e}_{2_{k,l}}$, and edges $\textbf{e}_{3_{m}}$ to Strasdat's pose graph optimization. Minimizing $\textbf{e}_{1_{i, j}}$ and $\textbf{e}_{2_{k,l}}$ suppresses changes in the relative transformations between camera poses, with the exception of gradual scale changes. Minimizing $\textbf{e}_{3_{m}}$ keeps the positions of the geo-tagged images close to the positions obtained from the associated geo-tags. Our overall cost function $E_{PGO}$ is defined as follows:
\begin{equation}
\begin{split}
E_{PGO}(\bigl\{ \bvec{S}_{i} \bigl\}_{i \in C_{1} \cup C_{2}}) 
&= \lambda_{1}\sum _{(i,j) \in C_{3}} \textbf{e}_{1_{i,j}}^{\mathrm{T}} \textbf{e}_{1_{i,j}} \\
&+
\lambda_{2}\sum _{(k,l) \in C_{4}} \textbf{e}_{2_{k,l}}^{\mathrm{T}} \textbf{e}_{2_{k,l}} +
\lambda_{3}\sum _{m \in C_{2}} \textbf{e}_{3_{m}}^{\mathrm{T}} \textbf{e}_{3_{m}}
\end{split}
\end{equation}
The corrected camera poses of keyframes $\bvec{S}_{n}$ and geo-tagged images $\bvec{S}_{m}$ are obtained by minimizing the cost function $E_{PGO}$ on Lie manifolds using the LM algorithm. Following this optimization, we also reflect this correction in the 3D map points, as in~\cite{strasdat2010scale}.

\subsection{Bundle Adjustment (BA) \label{sec:BA}}
Following the pose graph optimization, we refine the 3D reconstruction by applying bundle adjustment with the constraints of the geo-tagged images. Bundle adjustment is a classic method that jointly refines the 3D structure and camera poses (and camera intrinsic parameters) by minimizing the total re-projection errors. Each re-projection error $\bvec{r}_{i,j}$ between the $i^{th}$ 3D point and $j^{th}$ camera is defined as:
\begin{equation}\label{eq:ba1}
\bvec{r}_{i,j} = \bvec{x_{i}} - \pi(\bvec{R}_{j} \bvec{X}_{i} + \bvec{t}_{j})
\end{equation}
\begin{equation}\label{eq:ba2}
\pi(\bvec{p}) = [
		f_{x} \dfrac{\bvec{p}_{x}}{\bvec{p}_{z}} + c_{x},~
        f_{y} \dfrac{\bvec{p}_{y}}{\bvec{p}_{z}} + c_{y}
        ]^{\mathrm{T}}
\end{equation}
\noindent 
where $\bvec{X}_{i}$ is a 3D point and $\bvec{x}_{i}$ is the 2D observation of that 3D point; $\bvec{R}_{j}$ and $\bvec{t}_{j}$ are the rotation and translation of the $j^{th}$ camera pose, respectively; $\bvec{p} = [\bvec{p}_{x}, \bvec{p}_{y}, \bvec{p}_{z}]^{\mathrm{T}}$ is a 3D point; $\pi(\cdot): \mathbb{R}^3 \mapsto \mathbb{R}^2$ is the projection function; $(f_{x}, f_{y})$ is the focal length; and $(c_{x}, c_{y})$ is the center of projection.

To incorporate global position information of geo-tagged images with bundle adjustment, we 
add a penalty term corresponding to the constraint for a geo-tagged image~\cite{lhuillier2012incremental}. The total cost function with this constraint is given by:
\begin{equation}\label{eq:ba3}
\begin{split}
\MoveEqLeft
E_{BA}(\bigl\{ \bvec{X}_{i} \bigl\}_{i \in C_{5}}, \bigl\{ \bvec{T}_{j} \bigl\}_{j \in C_{1}}) = \sum _{(i,j) \in C_{\scalebox{0.7}{map-kf}}} \rho(\bvec{r}_{i,j}^{\mathrm{T}} \bvec{r}_{i,j}) + \lambda \sum _{m \in C_{3}} \|  \bvec{t}_{m} -  \bvec{y}_{m} \|^{2}
\end{split}
\end{equation}
\noindent
where $\bvec{T}$ is a camera pose of a keyframe represented as an element of $\sethree$, $\rho$ is the Huber robust cost function, $C_{5}$ consists of map points observed by keyframes in $C_{1}$, and $C_{1}$ and $C_{3}$ are defined in \Sref{sec:PGO}.
Both the positions of 3D points and the camera poses of keyframes are optimized by minimizing the cost function on Lie manifolds using the LM algorithm. This step can potentially correct the 3D map more precisely when it starts from a reasonably good 3D map.

\section{Experiments}
In this section, we evaluate the proposed method on the M\'{a}laga dataset~\cite{blanco2014malaga}, using geo-tagged images obtained from Google Street View. We also investigate the performance of pose graph optimization and bundle adjustment using the KITTI Dataset~\cite{geiger2012we}.

\subsection{Implementation}
We obtained geo-tagged images from Google Street View at intervals of 5~m within the area where the video was captured. We set the cost function weights to $\lambda_{1} = \lambda_{2} = 1.0 \times {10}^{5}$ and $\lambda_{3} = 1.0$, and we employed the g2o library~\cite{kummerle2011g} for the implementation of the pose graph optimization and bundle adjustment. 

\subsection{Performance of the Proposed Method \label{sec:eval_proposal}}
To verify the practical effectiveness of the proposed method, we evaluate it on the M\'{a}laga dataset using geo-tagged images obtained from Google Street View.

\begin{figure}[t]
	\centering
		\includegraphics[width=0.9\linewidth]{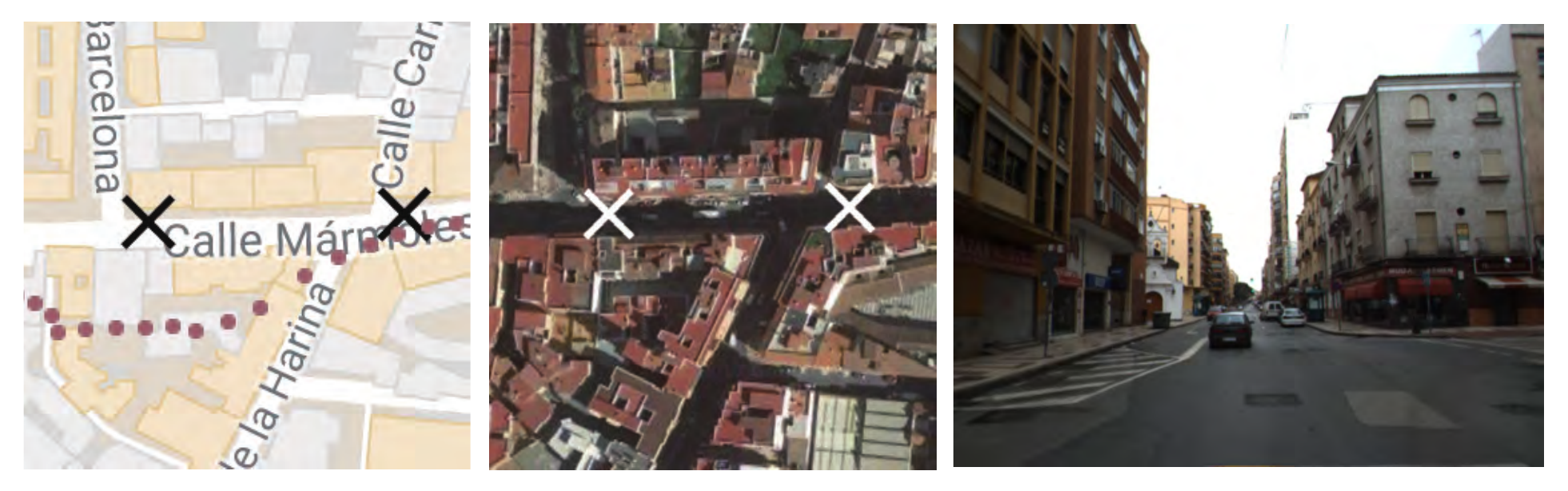}
		\caption{The left figure shows an example of inaccurate GPS data (brown dots) and manually assigned ground truth positions (back crosses) on Google Maps. Although we use Google Maps to visualize the results clearly, the shapes of roads are not sufficiently accurate. Our ground truth positions are always assigned in the appropriate lane of the road, as seen in the satellite image (white crosses in the center figure). The right figure shows an example of a video frame captured at the left of the two ground truth positions in the left figure.}
		\label{fig:gt_gps}
\end{figure}

The M\'{a}laga Stereo and Laser Urban Data Set (the M\'{a}laga dataset)~\cite{blanco2014malaga}---a large-scale video dataset that captures Street-View-usable areas---is employed in this experiment. The M\'{a}laga dataset contains a driving video captured at a resolution of 1024~$\times$~768 at 20~fps in a Spanish urban area. We extracted two video clips (video~1 and video~2) from the video, and used these for the evaluation. The two video clips contain no loops, and their trajectories are over 1~km long. All frames in the videos contain inaccurate GPS positions, which are sometimes confirmed to contain errors of more than 10~m. Because of the inaccuracies, we manually assigned the ground truth positions to some selected keyframes by referring to the videos, inaccurate GPS positions, and Google Street View 3D Map. \Fref{fig:gt_gps} presents an example of inaccurate GPS data and our assigned ground truth. Because the ground truth positions are assigned by taking into account the lane from which the video was taken, the errors in the ground truth are considered to be within 2~m, and these errors are sufficiently small for this experiment.

We evaluated the proposed method on the two videos by comparing the proposal and a baseline method that uses a similarity transformation (like a part of \cite{wang2013accurate}). For the baseline method, we apply the initialization (INIT: described in \Sref{sec:Init}) without applying pose graph optimization and bundle adjustment. We did not employ a global similarity transformation as a baseline because it cannot be applied until the end of the whole 3D reconstruction.

To evaluate the proposed method quantitatively, we considered the average (Ave) and standard deviation (SD) of 2D distances between the ground truth positions and corresponding keyframe positions in the UTM coordinate system (in meters). 

\begin{figure}[t]
	\begin{minipage}[t]{0.32\linewidth}
		 \centering
   			\includegraphics[width=0.95\linewidth]{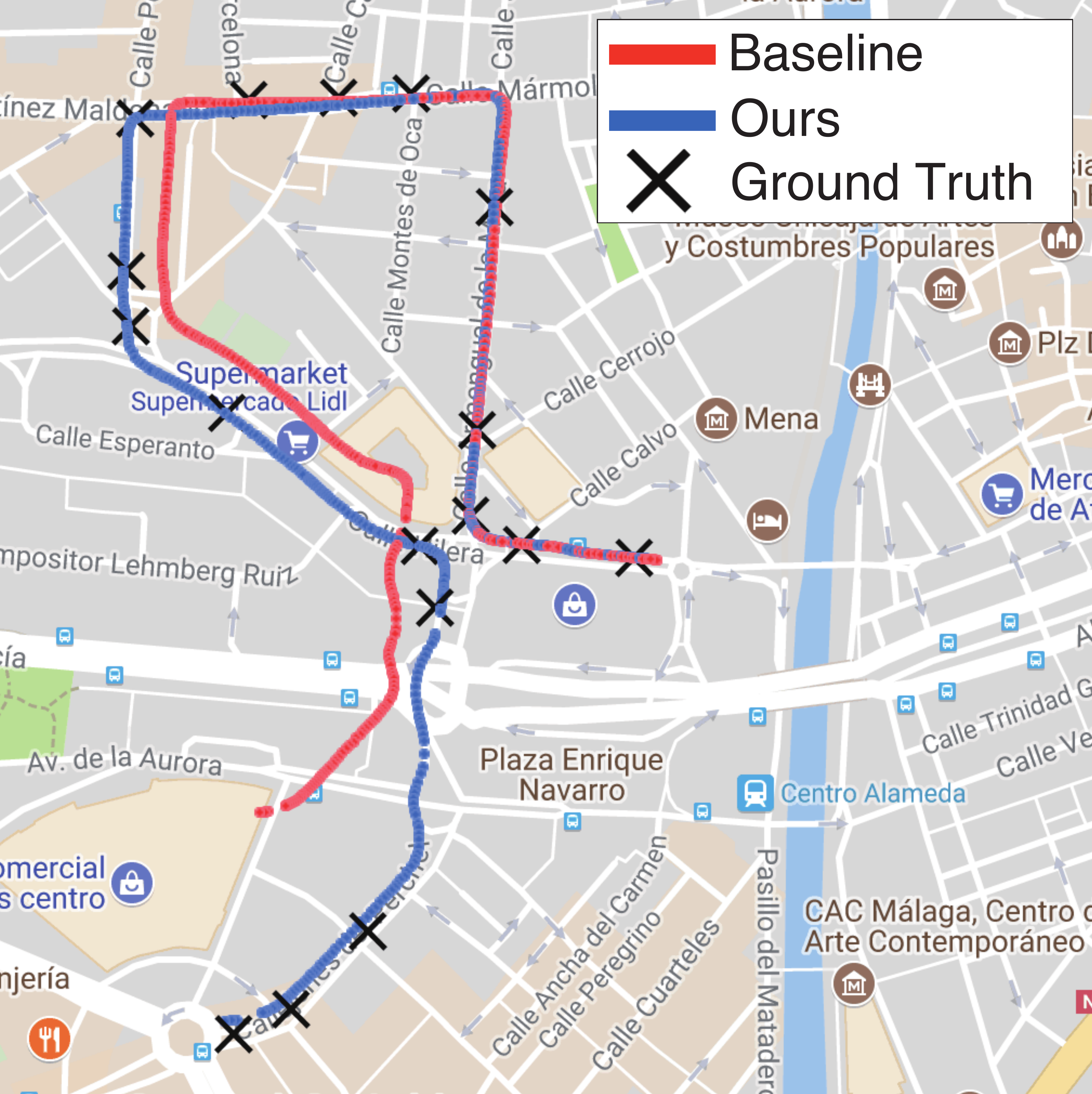}
  			 (a) camera trajectories
        \vspace*{0.2cm} 
	\end{minipage}
	\begin{minipage}[t]{0.32\linewidth}
  		\centering
   			\includegraphics[width=0.95\linewidth]{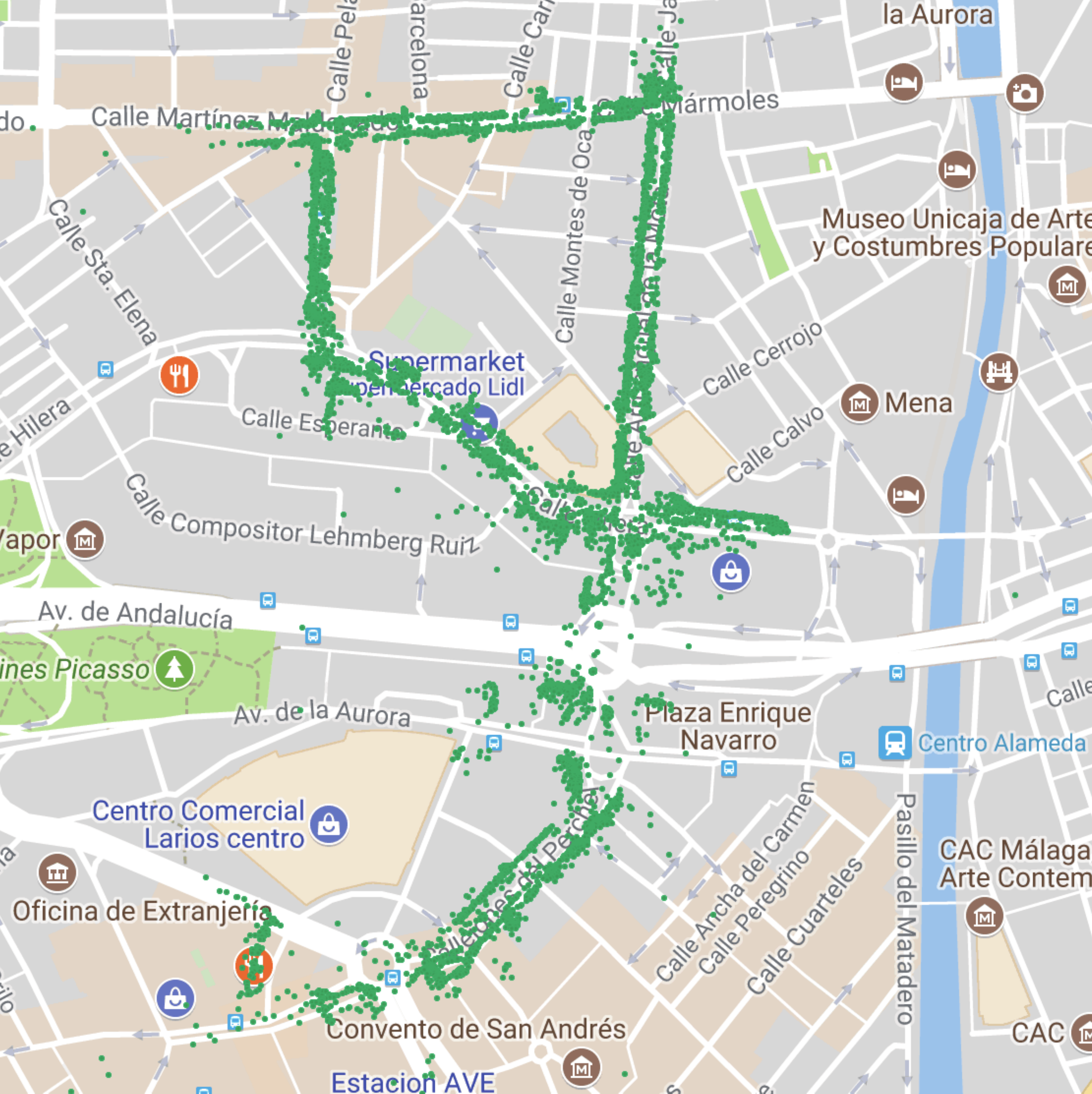}
  			(b) map points
        \vspace*{0.2cm}
	\end{minipage}
	\begin{minipage}[t]{0.32\linewidth}
  		\centering
   			\includegraphics[width=0.95\linewidth]{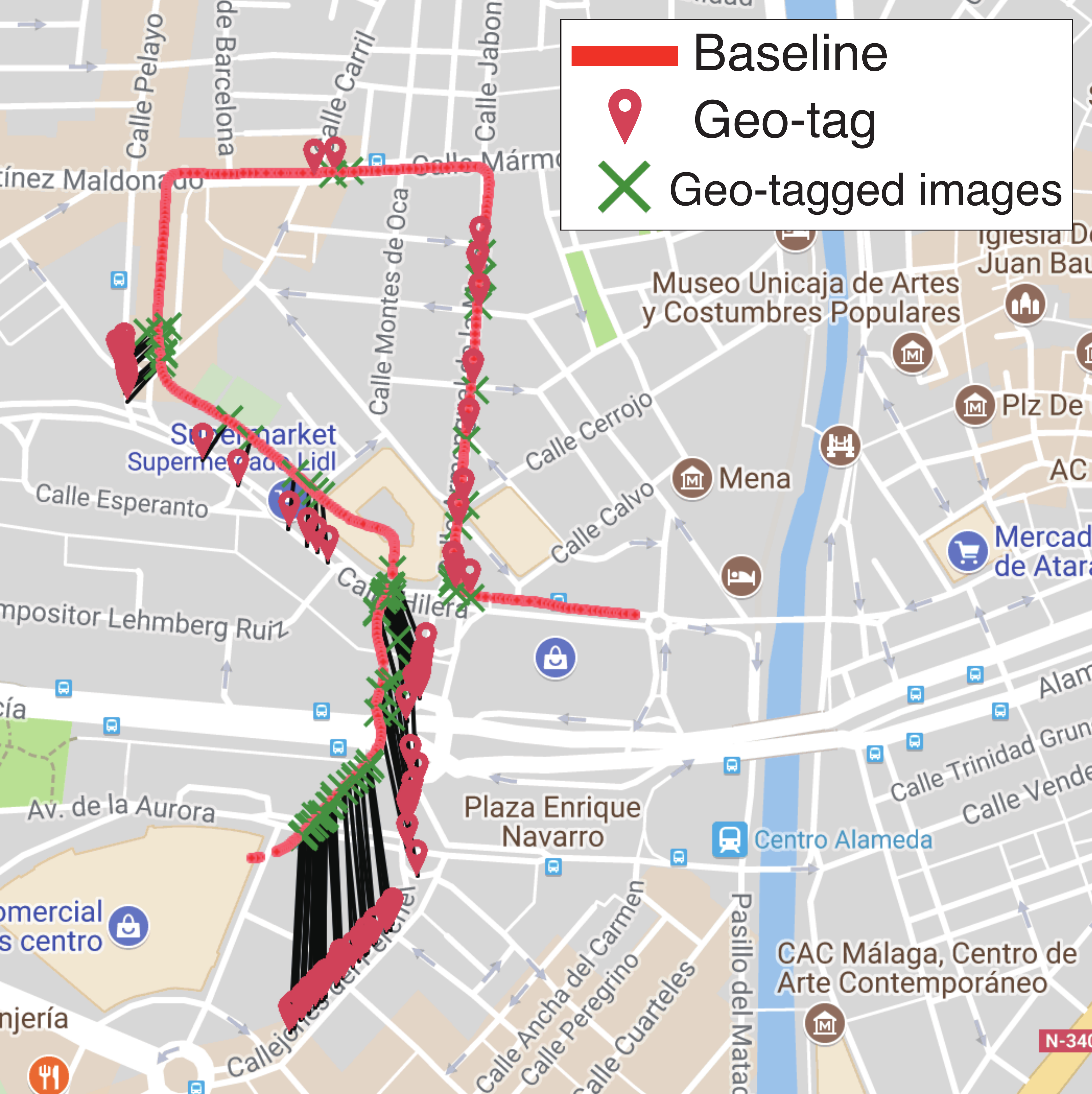}
  			(c) correspondences
        \vspace*{0.2cm}
	\end{minipage}
\\
    \begin{minipage}[t]{0.32\linewidth}
  		\centering
   			\includegraphics[width=0.95\linewidth]{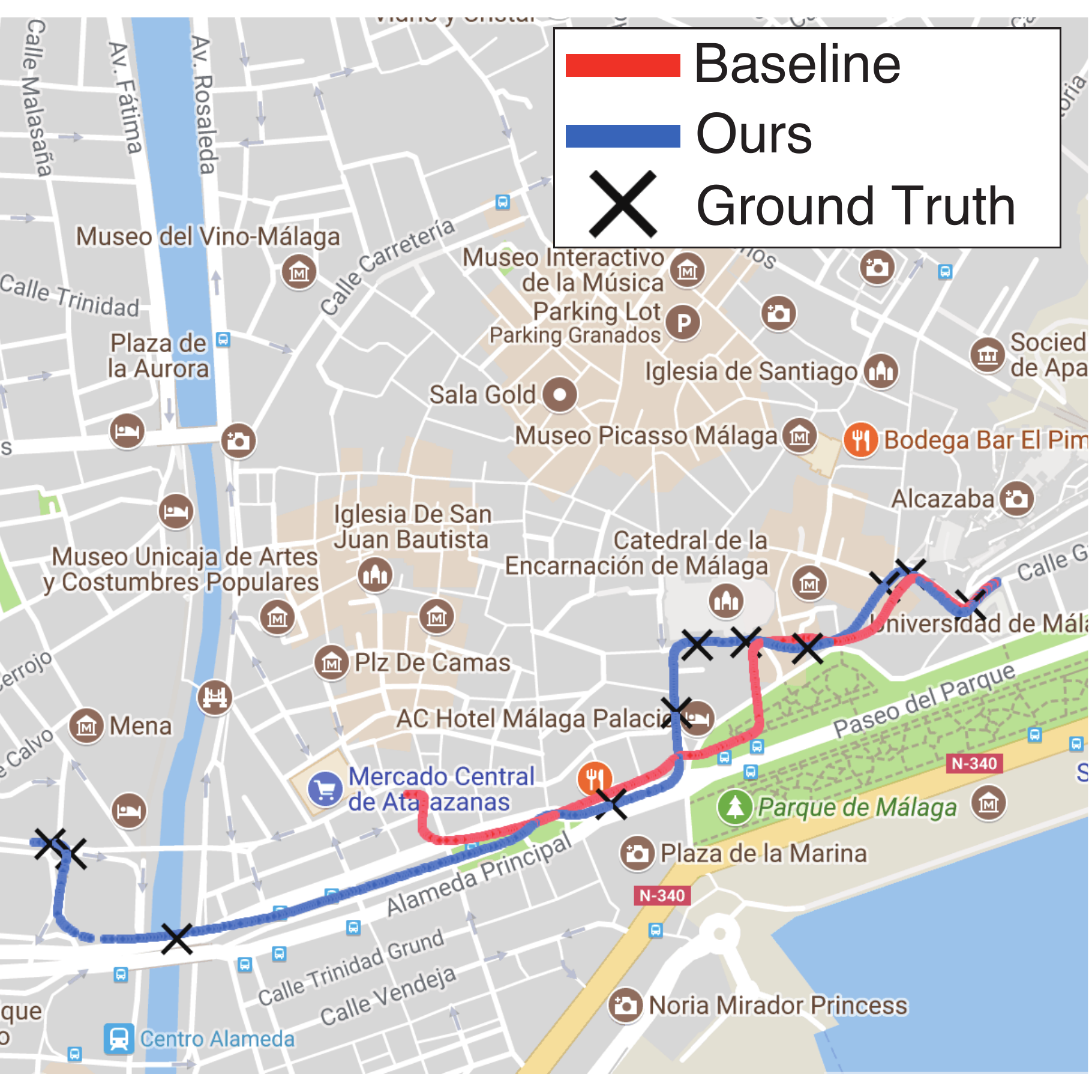}
  			(d) camera trajectories
	\end{minipage}
	\begin{minipage}[t]{0.32\linewidth}
  		\centering
   			\includegraphics[width=0.95\linewidth]{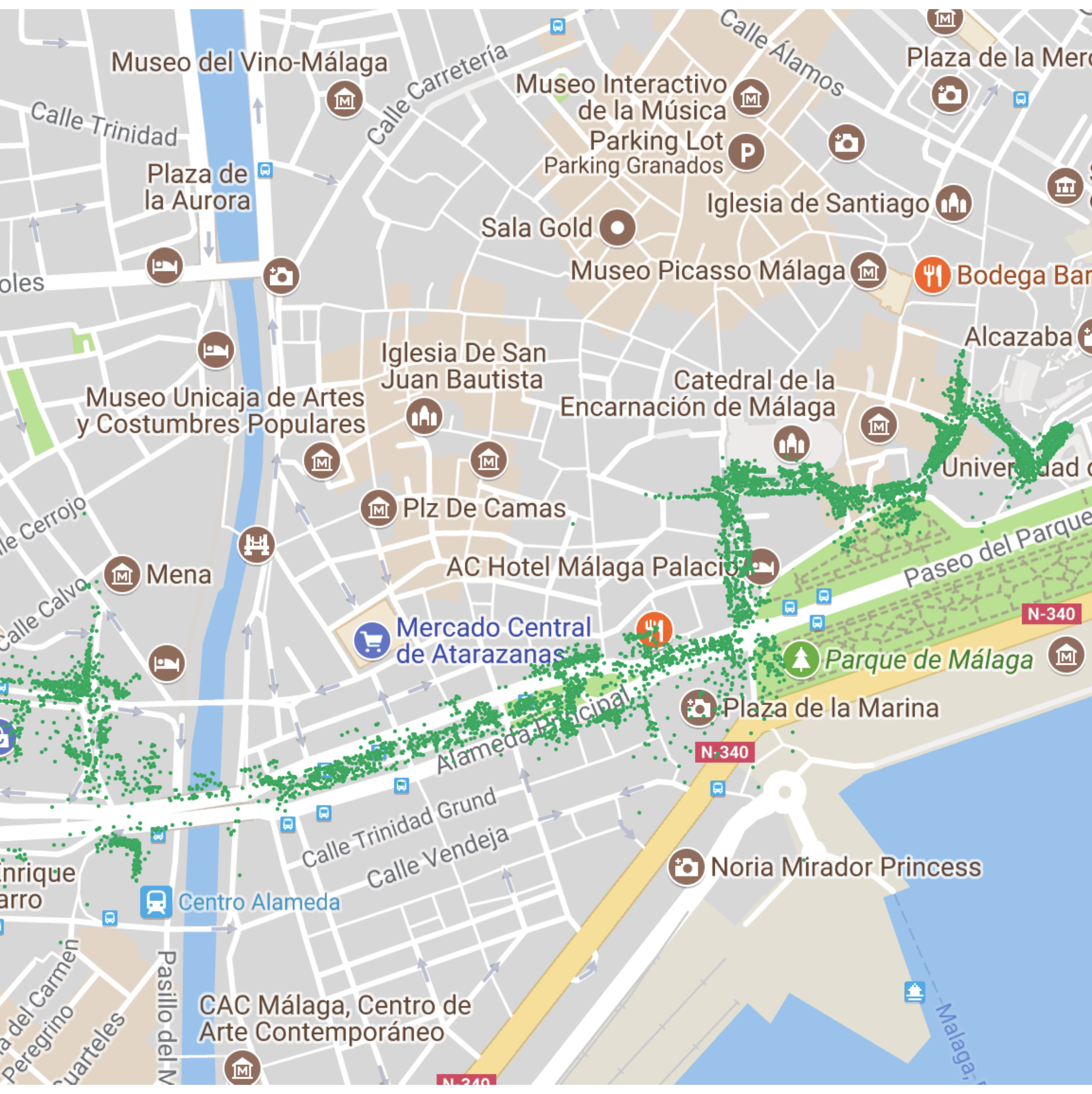}
  			(e) map points
        \vspace*{0.2cm}
	\end{minipage}
	\begin{minipage}[t]{0.32\linewidth}
  		\centering
   			\includegraphics[width=0.95\linewidth]{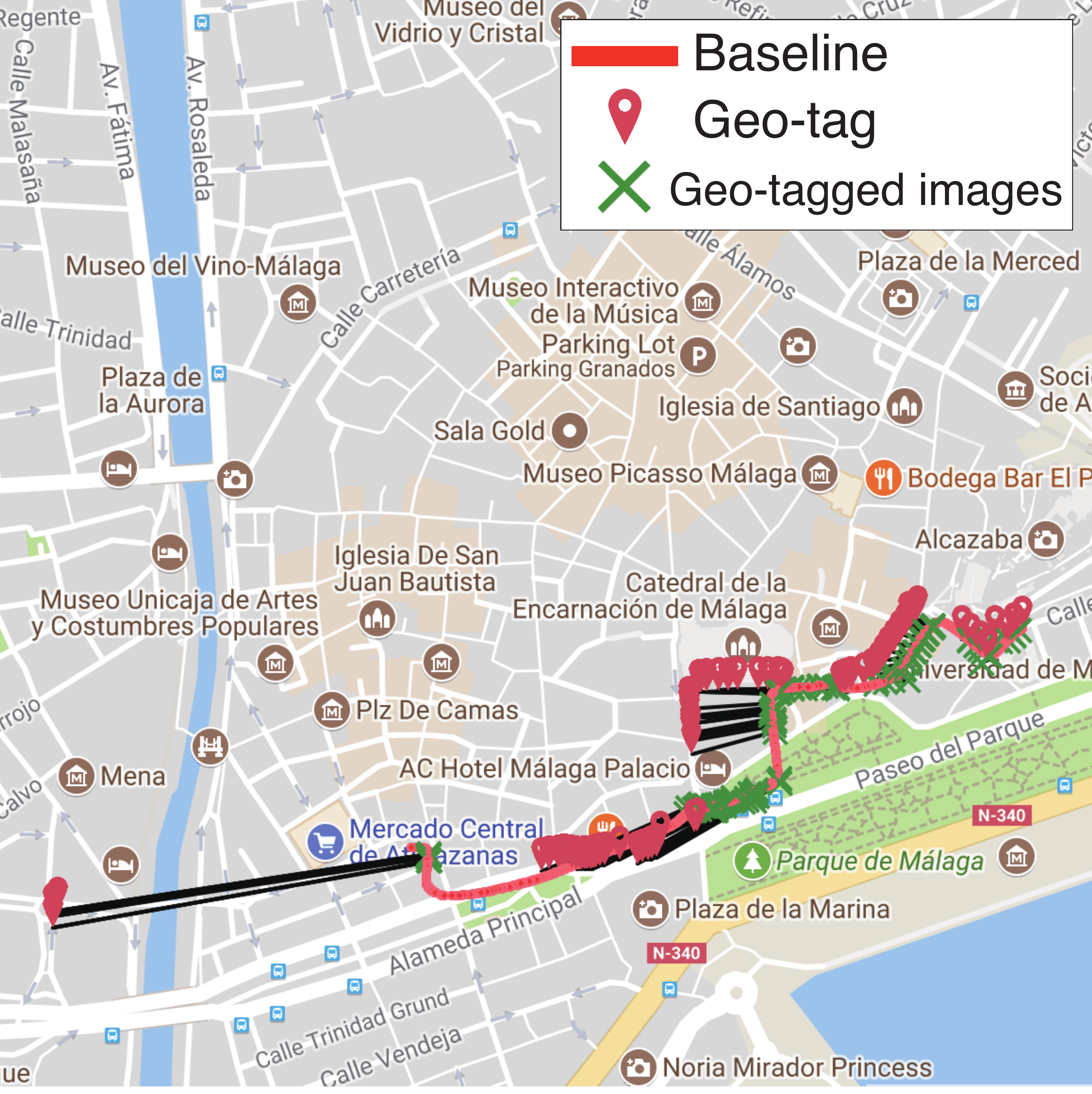}
  			(f) correspondences
	\end{minipage}
    \caption{Results of our proposed method visualized on Google Maps. Top: results on video~1. Bottom: results on video~2. In (a) and (d),  red and blue dots---which appear like lines---indicate the positions of keyframes corrected using a global similarity transformation (INIT) and our proposed method (Ours), respectively.  In (b) and (e), 3D map points corrected by our method are depicted by green dots. (c) and (f) show all of the employed correspondences between the positions of geo-tagged images transformed using a global similarity transformation (green crosses) and the positions of the corresponding geo-tags (red pin icons). The correspondences are applied incrementally for scale drift correction in our proposed method. }
    \label{fig:evaluation}
\end{figure}

\Tref{table:evaluation} presents the quantitative results, and \Fref{fig:evaluation} visualizes the results on Google Maps. As is clearly shown in these results, the baseline results accumulate scale errors, resulting in large errors of over 50~m. This is because the trajectories of these videos are long (greater than 1 km) and contain no loops. The proposed method sufficiently corrects scale drift, and significantly improves the 3D map by using geo-tagged images. In (b) and (e) of the visualized results, the 3D map points corrected using the proposed method are projected onto Google Maps, and it is shown that the 3D map points are correctly aligned to the map. To visualize all the correspondences between the 3D map coordinate system and the world coordinate system used in the proposal, we present the correspondences between the positions of geo-tagged images transformed by initialization and the positions of the corresponding geo-tags. These correspondences are employed incrementally for the correction. 

\begin{table}[t]
	\centering
		\caption{Results of our proposed method on the M\'{a}laga dataset using Google Street View.}
		\label{table:evaluation}
		\begin{tabular}{lSSSS}
		\toprule
		{} & \multicolumn{2}{c}{video~1} &  \multicolumn{2}{c}{video~2}\\
		\cmidrule(lr){2-3} \cmidrule(lr){4-5}
		 {} & {Ave [m]} & {SD} & {Ave [m]} & {SD}\\
		\midrule
		Baseline (INIT) & 54.8 & 141.3 & 142.5 & 249.8 \\
		Ours & 6.7 & 5.6 & 6.0 & 3.0 \\
		\bottomrule
		\end{tabular}
	
\end{table}

\subsection{Performance of PGO and BA}

To investigate the performance of the pose graph optimization and the bundle adjustment in our proposed method, we evaluated the performance using different combinations of these when varying the interval of $C_{\scalebox{0.7}{map-world}}$.

Through the previous experiment, we found that the geo-tag location information of Google Street View and the manually assigned ground truths of the M\'{a}laga dataset occasionally had errors of several meters. In this experiment, we control the interval of $C_{\scalebox{0.7}{map-world}}$, and use high-accuracy ground truths and geo-tags by using the KITTI dataset. The odometry benchmark of KITTI dataset~\cite{geiger2012we} contains 11 sequences of stereo videos and precise location information obtained from RTK-GPS/IMU, and unfortunately Google Street View is not available in Germany where this dataset was captured. The experiment was conducted on two sequences, which include the largest and second-largest errors when applying ORB-SLAM: sequences 02 and 08 (containing 4660 and 4047 frames, respectively). The left images of the stereo videos are used as input, and pairs of a right image and location information are identified as geo-tagged images. All the location information associated with keyframes is used as the ground truth. In this experiment with KITTI dataset, we can compare the performances of correction methods accurately for the following reasons: geo-tag information and ground truths are sufficiently precise (open sky localization errors of RTK-GPS/IMU $<$ 5 cm); and errors in geo-tagged image localization are sufficiently small, because keypoint matching between corresponding left and right images performs very well.

For the comparison, we present the results of the methods employing the initialization + the pose graph optimization (INIT+PGO), and initialization + the bundle adjustment (INIT+BA). The correction method of INIT + BA is the same as \cite{lhuillier2012incremental}, which is often used with a GPS location information. Ours includes the initialization, the pose graph optimization and the bundle adjustment. We changed the interval of geo-tagged images from 100 frames to 500 frames. For an equal initialization, we set geo-tagged images in the interval of 50 frames from the first to the $200^{th}$ frame.

\begin{table}[t]
	\renewcommand{\arraystretch}{1.2} 
	\centering
		\caption{Results of the experiments on the KITTI dataset: sequences 02 and 08. Values denote average 2D errors between ground truth positions and the corresponding keyframe positions [m].  Ours consists of INIT, PGO, and BA.}
		\label{table:kitti}
		\begin{tabular}{@{}lSSSSScSSSSS@{}}
		\toprule
		& \multicolumn{5}{c}{geotag interval (\#02)} &  \phantom{ab} &  \multicolumn{5}{c}{geotag interval (\#08)}\\
		\cmidrule(lr){2-6} \cmidrule(lr){8-12}
		&  \multicolumn{1}{r}{100} & \multicolumn{1}{r}{200} &  \multicolumn{1}{r}{300} &  \multicolumn{1}{r}{400\phantom{a}} &  \multicolumn{1}{r}{500} &&  \multicolumn{1}{r}{100} &  \multicolumn{1}{r}{200} &  \multicolumn{1}{r}{300} &  \multicolumn{1}{r}{400\phantom{a}} &  \multicolumn{1}{r}{500}\\
		\midrule
		INIT + BA & 1.15 & 2.22 & 65.86 & 164.17 & 96.66 && 0.45 &1.24 & 18.43 & 148.17 & 52.40 \\
		INIT + PGO & 4.57 & 4.26 & 7.54 & 10.89 & 11.96 && 0.93 & 2.83 & 4.64 & 5.44 & 9.11 \\
		Ours & 2.27 & 2.51 & 4.87 & 6.89 & 12.35 && 0.50 & 2.06 & 2.84 & 4.19  & 6.38 \\
		\bottomrule
		\end{tabular}
\end{table}

\begin{figure}[t]
	\begin{minipage}{0.49\linewidth}
  		\centering
   			\includegraphics[width=\linewidth]{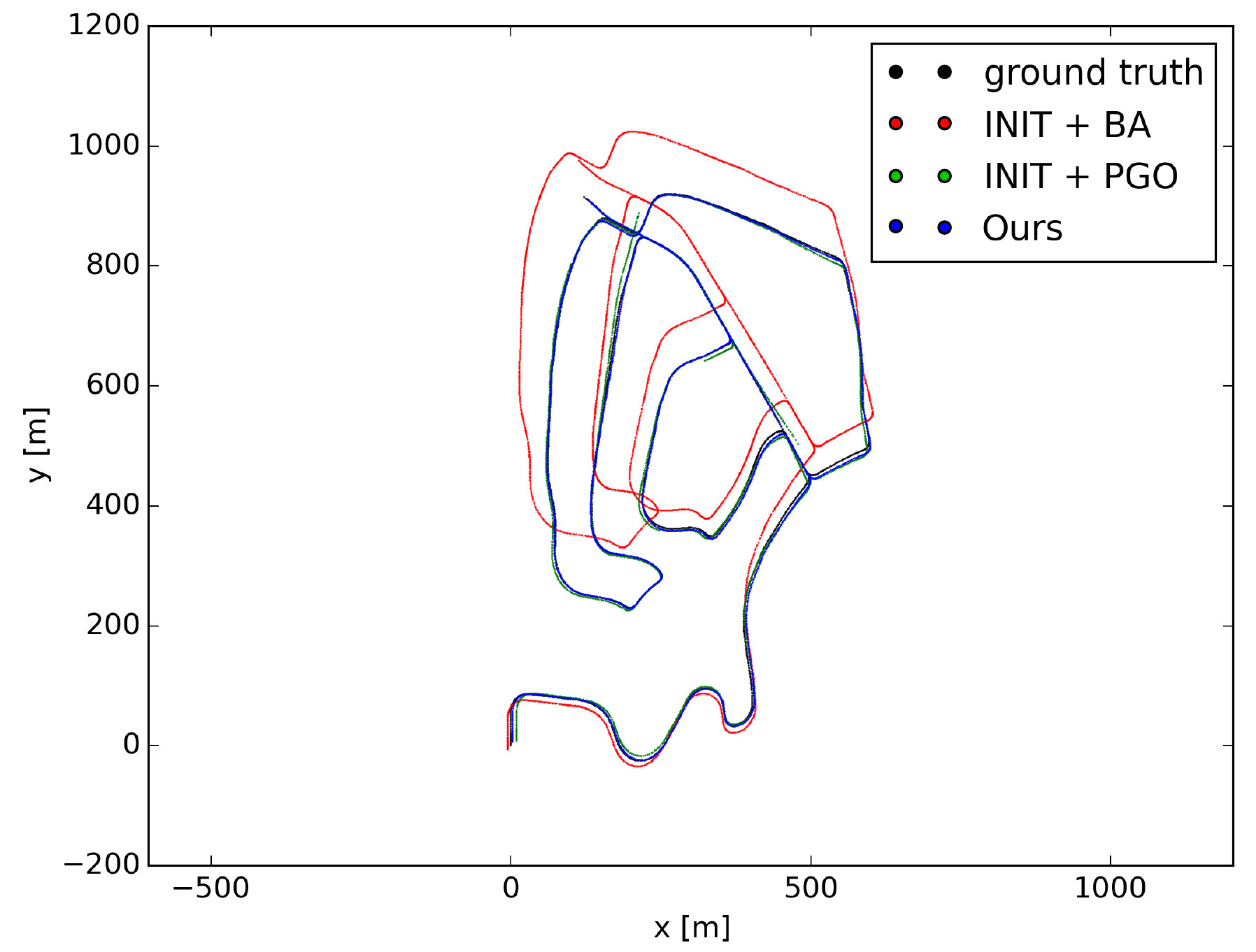}
  		\hspace{1.2cm} sequence 02
	\end{minipage}
	\begin{minipage}{0.49\linewidth}
  		\centering
   			\includegraphics[width=\linewidth]{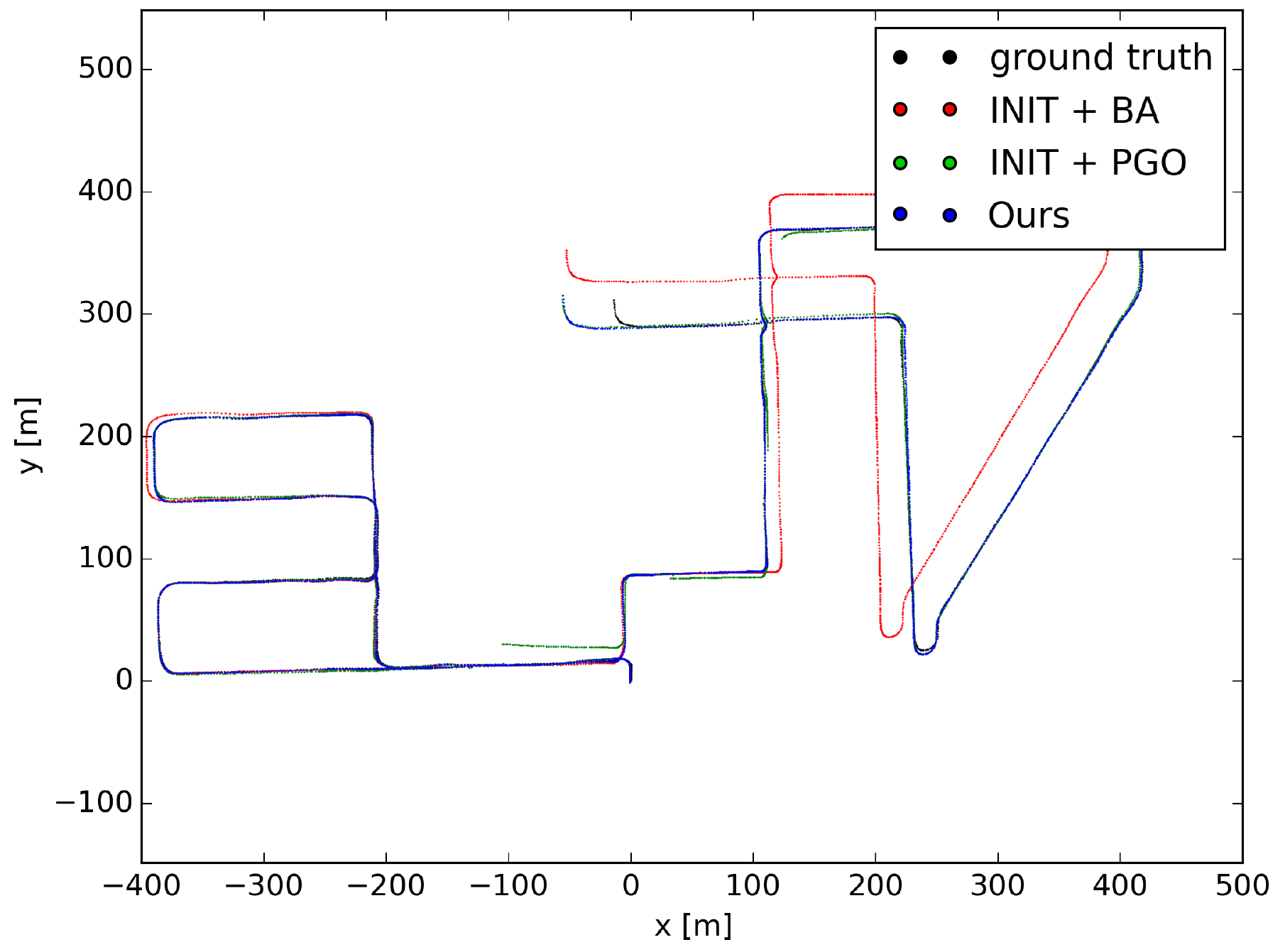}
  		\hspace{1.2cm} sequence 08
	\end{minipage}
	 \caption{Results of the experiment on the KITTI dataset when the interval of geo-tagged images is 300 frames. Keyframe trajectories estimated by INIT+BA, INIT+PGO, and Ours are visualized.}
	 
    \label{fig:kitti}
\end{figure}

\Fref{fig:kitti} visualizes the ground truth and keyframe trajectories estimated by INIT+BA, INIT+PGO, and Ours when the interval of geo-tagged images is 300 frames. \Tref{table:kitti} presents the quantitative results of the experiment, where the values represent the average 2D errors between ground truth positions and the corresponding keyframe positions in the UTM coordinate system (in meters). Moreover, we report the errors of the global linear transformation on the sequence 02 and 08 by aligning the keyframe trajectory obtained by ORB-SLAM with ground truths through a similarity transformation: $20.15$ and $25.12$, respectively. The results show that bundle adjustment with geo-tag constraints, which is typically employed in the fusion of 3D reconstruction and GPS information~\cite{lhuillier2012incremental}, is not suitable when the interval of $C_{\scalebox{0.7}{map-world}}$ is large. It can also be seen that Ours (the combination of initialization, pose graph optimization, and bundle adjustment) often estimates the keyframe positions more accurately than any other method.

\subsection{Scale Drift Correction}
\begin{figure}[t]
	\begin{minipage}{0.49\linewidth}
  		\centering
   			\includegraphics[width=0.85\linewidth]{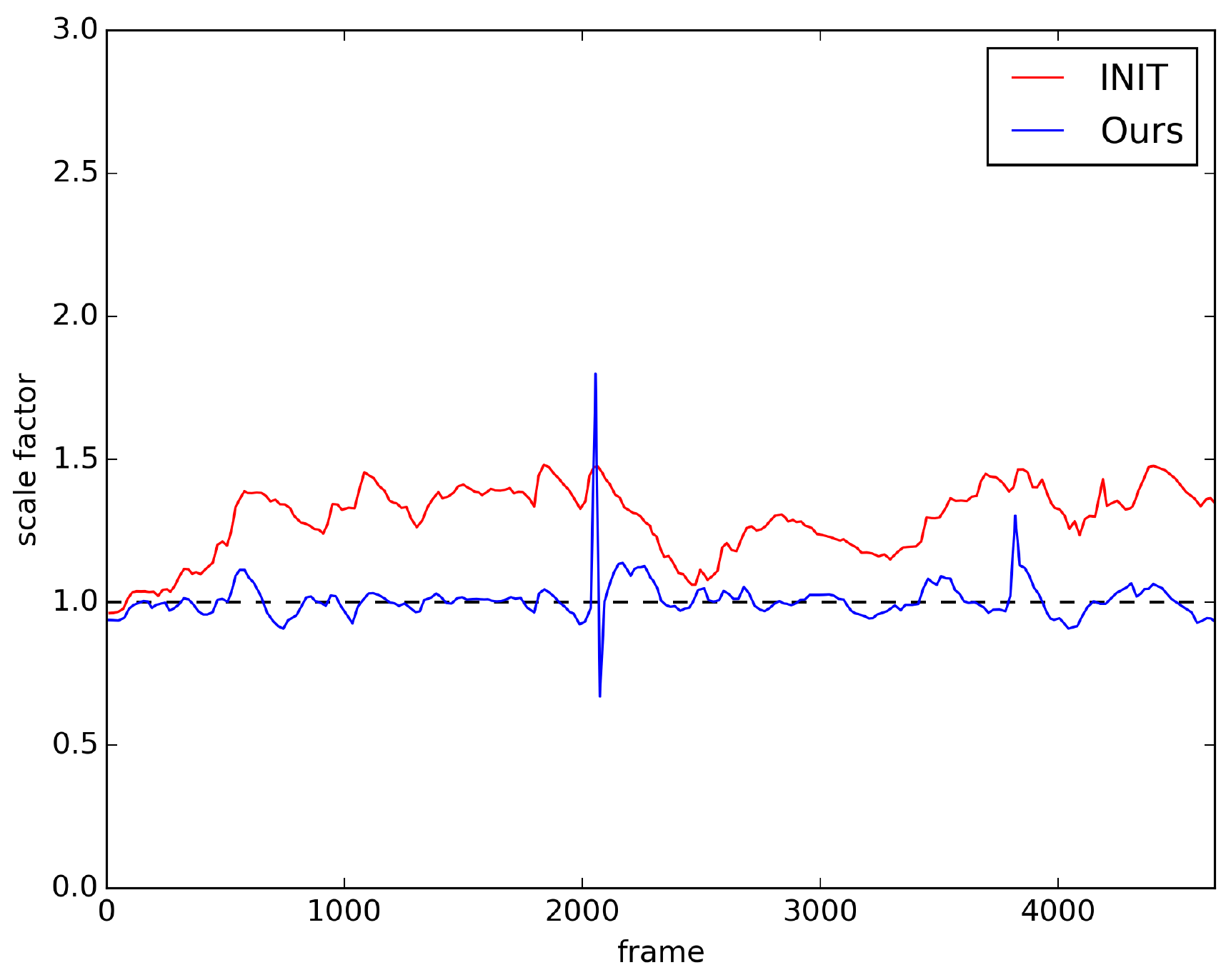}
  		\hspace{1.2cm} sequence 02
	\end{minipage}
	\begin{minipage}{0.49\linewidth}
  		\centering
   			\includegraphics[width=0.85\linewidth]{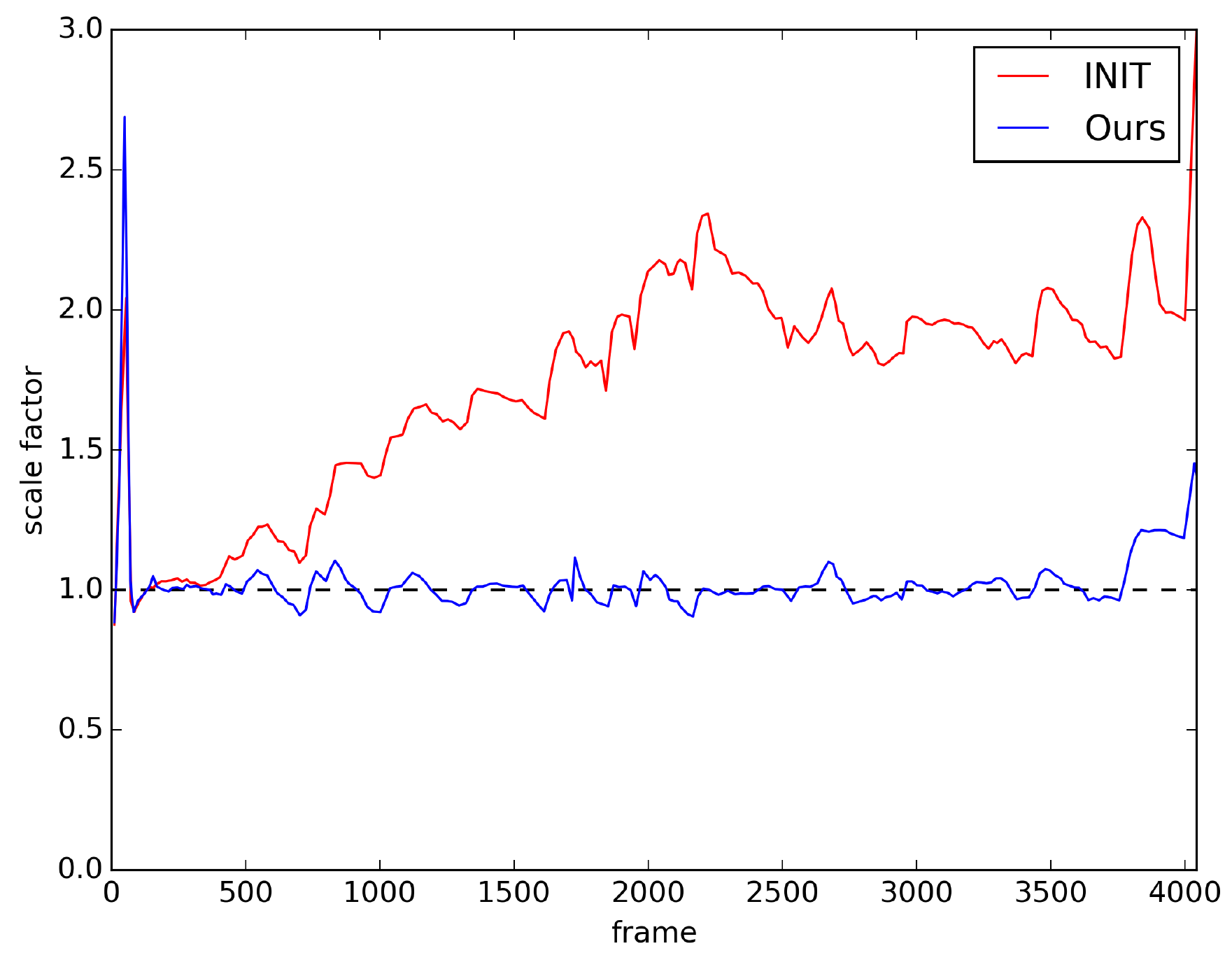}
  		\hspace{1.2cm} sequence 08
	\end{minipage}
	 \caption{Change in scale factor of the proposed method on the KITTI dataset sequences 02 and 08.}
    \label{fig:scale_change}
\end{figure}

To confirm that scale drift is corrected incrementally, we visualize the change in scale factor of the proposed method on the KITTI dataset sequences 02 and 08. \Fref{fig:scale_change} shows that ORB-SLAM with the initialization accumulates scale errors, and our method can keep the scale factor around $1$.

\section{Conclusion}
In this paper, we propose a novel framework for camera geo-localization that can correct scale drift by utilizing massive public repositories of geo-tagged images, such as those provided by Google Street View. By virtue of the expansion of such repositories, this framework can be applied in many countries around the world, without requiring the user to observe an environment. The framework integrates incremental SfM and a scale drift correction method utilizing geo-tagged images. In the correction method, we first acquire sparse 6-DoF correspondences between the 3D map coordinate system and the world coordinate system by using geo-tagged images. Then, we apply pose graph optimization over $\simthree$ constraints and bundle adjustment. Our experiments on large-scale datasets show that the proposed framework sufficiently improves the 3D map by using geo-tagged images. 

Note that our framework not only corrects the scale drift of 3D reconstruction, but also accurately geo-localizes a video. Our results are no less accurate than those of mobile devices (between 5 and 8.5~m) that use a cellular network and low-cost GPS~\cite{zandbergen2011positional}, and those using monocular video and road network maps~\cite{brubaker2016map} (8.1~m in the KITTI sequence 02 and 45~m in sequence 08). This implies that geo-localization using geo-tagged images is sufficiently useful compared with methods using other GIS information.

\section*{Acknowledgment}
This work was partially supported by VTEC laboratories Inc.


\bibliographystyle{splncs}
\bibliography{egbib}
\end{document}